\documentclass[sigconf]{acmart}
\usepackage{subfigure}
\usepackage{comment}
\usepackage{booktabs}
\usepackage{multirow}
\usepackage{amsmath}
\usepackage{latexsym}
\usepackage{algorithm}
\usepackage{algpseudocode}

\AtBeginDocument{%
  \providecommand\BibTeX{{%
    \normalfont B\kern-0.5em{\scshape i\kern-0.25em b}\kern-0.8em\TeX}}}

\setcopyright{rightsretained}
\copyrightyear{2022}
\acmYear{2022}
\newcommand{\model}{\textbf{MoleHD}}
\newcommand{\mypara}[1]{
	\vspace*{0.01cm}
	\noindent\textbf{\textit{#1}}}

\begin{document}

\title{MoleHD: Ultra-Low-Cost Drug Discovery using Hyperdimensional Computing}

\author{Dongning Ma,
    Rahul Thapa,
    Xun Jiao}
\email{{dma2, rthapa, xun.jiao}@villanova.edu}
\affiliation{%
  \institution{Villanova Univeristy}
  \state{Villanova, PA}
  \country{USA}
}

\begin{abstract}
Modern drug discovery is often time-consuming, complex and cost-ineffective due to the large volume of molecular data and complicated molecular properties. Recently, machine learning algorithms have shown promising results in virtual screening of automated drug discovery by predicting molecular properties. While emerging learning methods such as graph neural networks and recurrent neural networks exhibit high accuracy, they are also notoriously computation-intensive and memory-intensive with operations such as feature embeddings or deep convolutions. In this paper, we propose \model, an ultra-low-cost learning model based on brain-inspired hyperdimensional computing (HDC) for molecular property prediction. We develop HDC encoders to project SMILES representation of a molecule into high-dimensional vectors that are used for HDC training and inference. We perform an extensive evaluation using 29 classification tasks from 3 widely-used molecule datasets (Clintox, BBBP, SIDER) under three splits methods (random, scaffold, and stratified). By an comprehensive comparison with 8 existing learning models including SOTA graph neural networks (GNNs), we show that \model ~is able to achieve highest ROC-AUC score on random and scaffold splits on average across 3 datasets and achieve second-highest on stratified split. More importantly, \model ~achieves such performance with significantly reduced computing cost than GNNs, e.g., ~10 mins training using CPU VS. 5 days on GPU, and 80 KB VS 100MB model size. 
The promising results presented in this paper can potentially lead to a novel tiny ML paradigm in drug discovery research. 
\end{abstract}


\maketitle

\section{Introduction}
Drug discovery is the process of using multi-disciplinary knowledge such as biology, chemistry and pharmacology to discover proficient medications amongst candidates according to safety and efficacy requirements. Modern drug discovery often features a cost-ineffective virtual screening process to select candidates from general chemical databases such as \textit{ChEMBL}~\cite{gaulton2012chembl} and \textit{OpenChem}~\cite{kim2016pubchem} with large volume of molecular data to build a significant smaller in-house database for further synthesis. 

Traditional machine learning algorithms such as random forest~\cite{jayaraj2016gpurfscreen}, support vector machine~\cite{liew2009svm}, k nearest neighbors~\cite{arian2020protein}, and gradient boosting~\cite{wu2018moleculenet} have been investigated in drug discovery applications. Such algorithms use molecular representations as input to predict molecular properties. However, because of limited sophistication, deep and complex structural information within a molecule is generally overlooked by those models. Thus, they typically do not exhibit strong capability in learning the features and only achieve sub-par performance. On the other hand, inspired by the recent success from other applications such as computer vision, neural network models have been increasingly applied in drug discovery. GNN learns representations by aggregating nodes and neighbouring information for molecular property predictions under different drug discovery objectives. However, molecular graphs often requires pre-processing or featurization. Extended-connectivity fingerprints (ECFP) is one of the most common featurization method that converts molecular graphs into fixed length representations, or fingerprints~\cite{rogers2010extended}. Such featurization algorithms usually requires comprehensive efforts using chemical tool-chains such as RDKit~\cite{landrum2013rdkit}. 

This paper takes a radical departure from common machine learning methods including neural networks by developing an ultra-low-cost brain-inspired hyperdimensional computing (HDC) model that requires less pre-processing efforts and is easier to implement. Inspired by the attributes of brain circuits including high-dimensionality and fully distributed holographic representation, this emerging computing paradigm postulates the generation, manipulation, and comparison of symbols represented by high-dimensional vectors. Compared with DNNs, the advantages of HDC include smaller model size, less computation cost, and one/few-shot learning, making it a promising alternative computing paradigm~\cite{karunaratne2020memory}. Recently, HDC has demonstrated success on various application domains such as robotics~\cite{mitrokhin2019learning}, natural language processing~\cite{thapa2021spamhd}, biomedical signal analysis~\cite{rahimi2016hyperdimensional}, and biological sequence matching~\cite{imani2018hdna}.

In this paper, we develop \model, an HDC-based method to predict molecular properties in drug discovery. \model ~first tokenizes SMILES strings into numerical list of tokens, and then develop HDC encoding mechanisms to project realistic features into their high-dimensional space representations: hypervectors. Next, \model ~leverages hypervector properties to train an HDC model that can be used to perform molecule classification tasks.

The qualitative advantages of \model ~compared to existing neural network-based classifiers for drug discovery are: (1) back-propagation free: \model ~does not need backpropagation to train a set of parameters; instead, it uses one/few-shot learning to establish abstract patterns that can represent specific symbols. (2) efficient computing: unlike neural networks, \model ~does not need complicated arithmetic operations such as convolutions which presents a major computing/energy burden to computing platforms; instead, it only uses simple arithmetic operations such as addition between two vectors. Thus, \model ~only needs to run on commodity CPU and can finish both training and testing on the reported datasets within minutes, while GNN requires around 5 days for training using Nvidia GPU~\cite{wang2021molclr}. (3) smaller model size: \model ~only needs to store a set of vectors for comparison during inference, while SOTA neural networks often need millions of parameters and requires memory in 100MB scale to store the parameters (e.g., weights and activation values)~\cite{ma2020multi}. 
The main contributions of this paper are summarized as below:
\begin{enumerate}
    \item We propose \model, an ultra-low-cost novel learning model based on hyperdimensional computing. This promising results of \model ~provide a viable option and alternative to existing learning methods in drug discovery domain. 
    \item We develop a complete molecular-specific pipeline for HDC-based drug discovery. First, \model ~tokenizes SMILE strings into tokens representing the substructures and then project them into hypervectors during encoding. Then, \model ~uses the encoded hypervectors to train and evaluate the classification model. 
    \item We perform an extensive evaluation of \model ~on 29 classification tasks from 3 widely-used molecule datasets under three splits methods. By a comprehensive comparison with 8 baseline models including SOTA neural networks, \model ~is able to achieve highest ROC-AUC score on random and scaffold splits on average across 3 datasets and achieve second-highest on stratified split. More importantly, \model ~achieves such performance with significantly reduced computing cost than GNNs, e.g., ~10 mins training using CPU VS. 5 days on GPU, and 80 KB VS 100MB model size. 
    \item We conduct a design space exploration of \model ~by developing two tokenization schemes (\model-PE and \model-char), two gram sizes (uni-gram and bi-gram), and evaluate their corresponding performance. 
\end{enumerate}

\section{Related Works}
Hyperdimensional computing (HDC), also known as vector-symbolic architectures (VSA), was introduced as an alternative computational model mimicking the ``human brain'' at the functionality level~\cite{kanerva2009hyperdimensional}.
HDC has been used in modern robotics to perform active perception by integrating the sensory perceptions experienced by an agent with its motoric capabilities, which is vital to autonomous learning agents~\cite{mitrokhin2019learning}. HDC has also been used in biomedical signal processing and exhibits 97.8\% accuracy on hand gesture recognition based on EMG, which surpasses support vector machine by 8.1\%~\cite{rahimi2016hyperdimensional}. Recent works also show that HDC outperforms other machine learning methods in DNA sequencing~\cite{imani2018hdna, kim2020geniehd}.

Machine learning algorithms are used in drug discovery mostly in predicting molecular properties to determine if they satisfy the drug discovery objective. 

Recently, emerging machine learning algorithms such as GNNs are increasingly applied to drug discovery for achieving higher performance. GNNs leverages fingerprints derived from the molecular graph to learn the representations. Direct message passing neural network (D-MPNN) is an evolution of message passing neural networks that centers on bonds between atoms which is able to maintain two representations~\cite{yang2019learned, swanson2019message}. Contrastive learning is also applied into GNNs to fuse drug discovery domain knowledge and molecular properties to augment learning of representations~\cite{fang2021knowledge, wang2021molclr}. In addition to GNNs, natural language processing (NLP) models such as recurrent neural networks (RNNs) are also introduced in drug discovery. Compared with GNNs, RNNs typically do not rely on complex fingerprint conversion process using toolchains such as \textbf{RDKit}~\cite{quan2018system, lin2020novel}. However, RNNs still require word embeddings tools such as \textbf{Smi2Vec}, to fully extract features from the molecule SMILES representation.

\section{Preliminaries on HDC}

\subsection{Hypervectors} Hypervectors (HV) are high-dimensional (usually higher than 10,000), holographic (not micro-coded) vectors with (pseudo-)random and i.i.d. elements~\cite{kanerva2009hyperdimensional}. An HV with $d$ dimensions can be denoted as $\overrightarrow{H} = \langle h_1, h_2, \dots, h_d\rangle$, where $h_i$ refers to the elements inside the HV. HVs are fundamental blocks in HDC that are able to accommodate and represent information in different scales and layers. When the dimensionality is sufficiently high (e.g., $D=10,000$), any two random HVs are nearly orthogonal~\cite{kanerva2009hyperdimensional}. HDC utilizes different operations HVs support as means of producing aggregations of information or creating representations of new information.

\subsection{Operations}
In HDC, addition ($+$), multiplication ($*$) and permutation ($\rho$) are the three basic operations HVs can support. Additions and multiplications take two input HVs as operands and perform \textbf{element-wise} add or multiply operations on the two HVs. Permutation takes one HV as the input operand and perform \textbf{cyclic rotation} by a specific amount. All the operations do not modify the dimensionality of the input HVs, i.e. the input and the output HVs are in the same dimension.

These three operations also have their corresponding physical meanings. Addition is used to aggregate same-type information, while multiplication is used to combine different types of information together to generate new information. Permutation is used to reflect spatial or temporal changes in the information, such as time series or spatial coordinates~\cite{kanerva2009hyperdimensional}.

\subsection{Similarity Measurement}

In HDC, the similarity metric $\delta$ between the information that two HVs represent is measured by similarity check. Different algorithms can be used to calculate the similarity, such as the Euclidean ($L2$) distance, the Hamming distance (for binary HVs), and cosine similarity (which we use in this paper). A higher similarity $\delta$ between two HVs shows that these two HVs have more information in common, or vice versa. Because of the high dimensionality of HVs, addition generally results in a new HV that is approximately 50\% similar to the two original HVs, while multiplication and permutation result in HVs that are orthogonal to the original HVs, i.e., not similar. 

\begin{figure*}
    \centering
    \includegraphics[page=1, width=1.93\columnwidth]{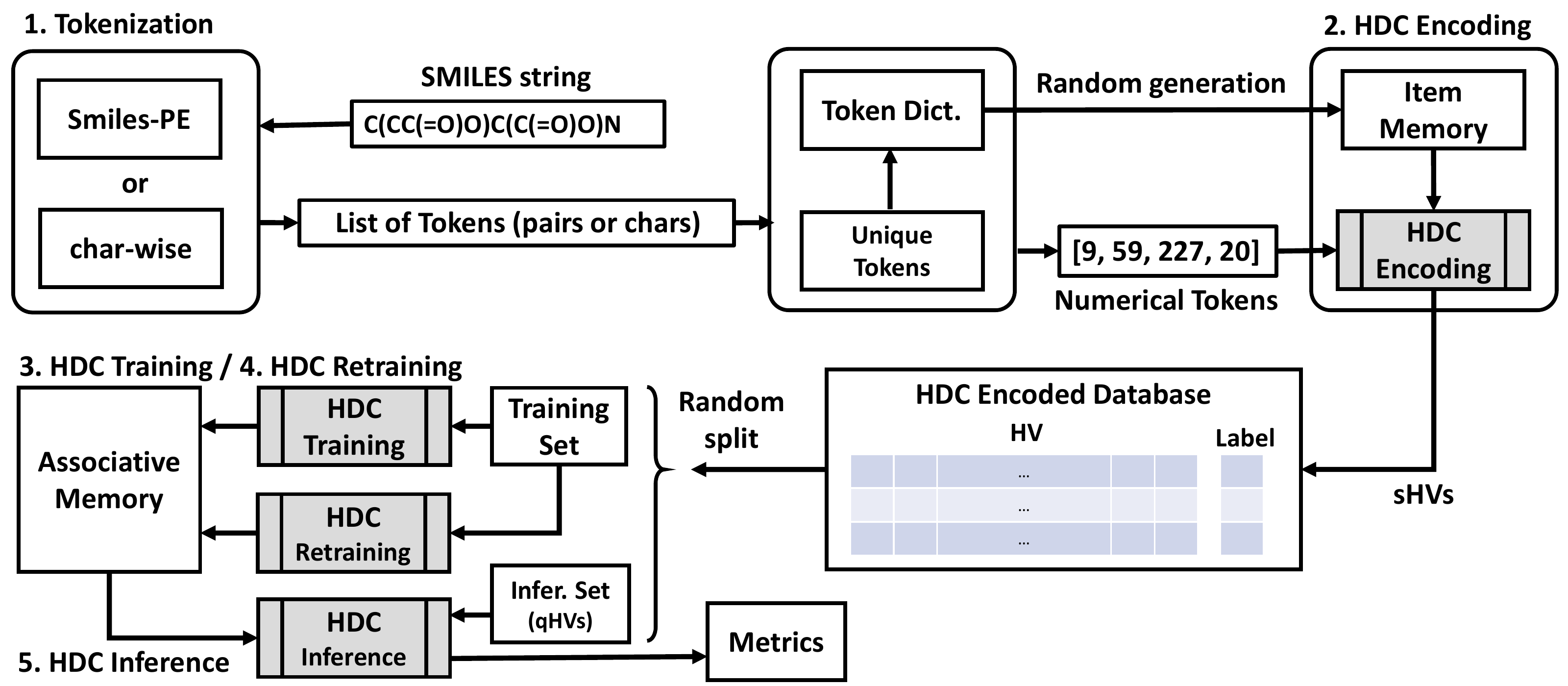}
    \caption{Overview of \model. \model ~has 5 major steps: \textbf{Tokenization}, \textbf{Encoding}, \textbf{Training}, \textbf{Retraining} and \textbf{Inference}.}
    \label{fig:moleculeHD}
\end{figure*}

\section{\model ~Framework}
In this section, we will introduce the proposed framework \model ~and how it utilizes HDC to perform learning tasks in drug discovery. An overview of \model ~is presented in Fig.~\ref{fig:moleculeHD}.

\subsection{Tokenization}
In \model, tokenization is the process of converting molecule features into their corresponding set of numerical tokens. It basically consists of three procedures: converting the SMILES string into a list of tokens and then assign number for the tokens to obtain a list of numerical tokens which are ready for HDC processing.

We develop two tokenization schemes for \model: \model-char and \model-PE. \model-char is the basic tokenization strategy that treats the input SMILES string as a textual string. \model-char split the textual string into characters to obtain a list of tokens. Each unique character inside the string is then assigned with a unique random number to form the numerical tokens. \model-PE uses the open-source SMILES Pair Encoding (SMILES-PE) model to extract the sub-structures in the input SMILES strings then assign a unique number based on their appearance frequency ranking to tokenize them. SMILES-PE is a data-driven algorithm to find substructures from a SMILES string~\cite{li2020smiles}. \model-PE ~uses SMILES-PE as-is as an add-on and does not require additional time for the pre-train. Due to model size limitation or other user-specific constraints, only $m$ tokens will be stored. For missing tokens in \model, a special token `0' is assigned.

\begin{figure*}
    \centering
    \subfigure[HDC Encoding]{
        \includegraphics[page=1, width=1.94\columnwidth]{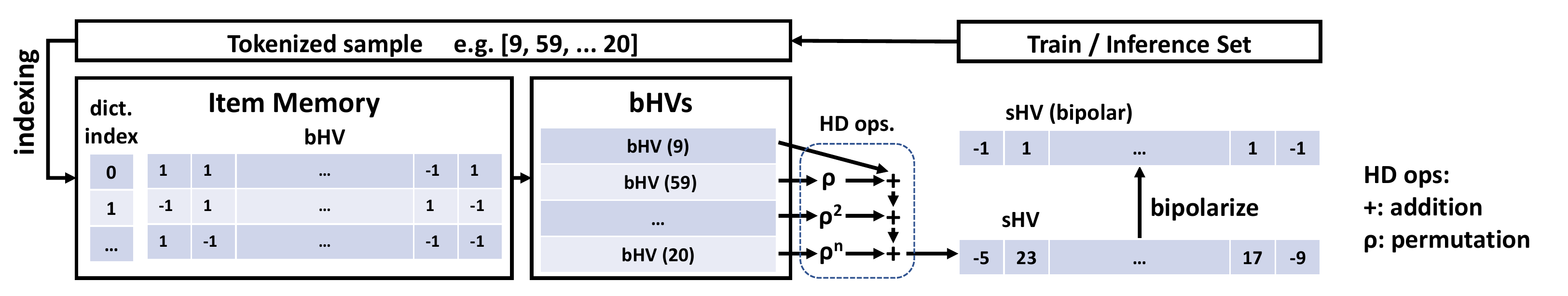}
    }
    \subfigure[HDC Training and Retraining]{
        \includegraphics[page=2, width=1.99\columnwidth]{./figures/hdc.pdf}
    }
    \subfigure[HDC Inference]{
        \includegraphics[page=3, width=1.94\columnwidth]{./figures/hdc.pdf}
    }
    \caption{HDC processing: \textbf{Encoding}, \textbf{Training}, \textbf{Retraining} and \textbf{Inference}.}
    \label{fig:hdc}
\end{figure*}
\subsection{HDC Encoding}
Encoding is the process to project real-world features into their high-dimensional space representations: the HVs. In \model, encoding process projects tokenized sample into its representing sample HV (sHV, or $\overrightarrow{S}$) via a combination of pre-defined HD operations as shown in Fig.~\ref{fig:hdc}(a).

\mypara{Item Memory}
Item memory is generated from the token dictionary in tokenization. The item memory contains base HVs (bHV, or $\overrightarrow{B}$) in the same number ($m+1$, considering the missing entry assigned as `0') as the entries in the token dictionary, i.e., each HV serves as the high-dimensional representation of a token. The item memory is fully random generated using a seed to ensure the i.i.d. properties. We note item memory as $\mathbb{B} = \{\overrightarrow{B_0}, \overrightarrow{B_1}, ... , \overrightarrow{B_m}\}$ where $\overrightarrow{B_i}$ is the base HV with index $i$.

\mypara{HD operations in Encoding}
In \model, encoding schemes can be flexible and data-specific. Algorithm~\ref{alg:unigram} shows the process of uni-gram encoding as an example. Tokenized sample first uses its tokens iteratively in the item memory to index and fetch the corresponding base HVs. The base HVs permutate by their order in the tokenized sample and added up to establish the sample HV (Line 2 - 4). \model ~also bipolarizes the elements inside the sample HV according to their relation with zero (Line 5 - 11). \model ~also features bi-gram and tri-gram encoding which resembles the uni-gram encoding but instead permutes every 2 or 3 tokens aggregated together by HV multiplication.

\begin{algorithm}
\small
    \caption{\model ~Encoding (uni-gram)}  
    \algrenewcommand\algorithmicrequire{\textbf{Input}}
    \algrenewcommand\algorithmicensure{\textbf{Output}}
    \begin{algorithmic}[1]
    \Require tokenized sample $T = \{t_0, t_1, t_k\}$, item memory $\mathbb{B}$.
    \Ensure sample HV $\overrightarrow{S}$
    \State $setZero(\overrightarrow{S})$ 
    \For{$t_i$ in $T$} \textit{\textbackslash * Perform uni-gram encoding. */}
        \State $\overrightarrow{S} = \overrightarrow{S} + \rho^i(\overrightarrow{B_{t_i}})$ 
    \EndFor
     \For{$s_i$ in $\overrightarrow{S}$} \textit{\textbackslash * Bipolarize the sample HV. */}
        \If{$s_i > 0$} $s_i = 1$
        \ElsIf{$s_i < 0$} $s_i = -1$
        \EndIf
     \EndFor
    \end{algorithmic}
    \label{alg:unigram}
\end{algorithm}

\subsection{HDC Training}
Training is the process of establishing the associative memory $\mathbb{C} = {\overrightarrow{C_1}, \overrightarrow{C_2}, ... , \overrightarrow{C_p}}$ using the training set. Associative memory (AM) contains $p$ class HVs (cHV, or $\overrightarrow{C}$), each representing a class in a learning task. Using a binary classification task as example shown in Fig.~\ref{fig:hdc}(b), AM contains the class HV representing positive ($\overrightarrow{C_P}$) and negative ($\overrightarrow{C_N}$). For each training sample, \model ~adds its HV to the corresponding class HV according to the label, as shown in Eq.~\ref{eqn:retrain}. This process is to aggregate the information from sample HVs together into the AM. However, one-epoch training is usually not enough to train a reliable AM for learning tasks, it is necessary to perform additional epochs for fine-tuning or retraining.

\begin{equation}
    \begin{aligned}
       \overrightarrow{C_P} = \sum \overrightarrow{S_p}, & & \overrightarrow{C_N} = \sum \overrightarrow{S_n}
    \end{aligned}
    \label{eqn:train}
\end{equation}

\subsection{HDC Retraining}
Retraining is the process of fine-tuning the associative memory to enhance its accuracy using the training set, as shown in Fig.~\ref{fig:hdc}(b). For each training sample, \model ~tries to use the AM to predict its label. If the prediction is correct, \model ~proceeds to the next training sample. However, if the prediction is wrong, it indicates that the correct information of the sample HV has not been aggregated into the AM, or the information in the AM is not properly represented. Therefore, \model ~performs an update to the AM to remove the erroneous and add the correct information, by subtracting the sample HV from the wrongly predicted class HV ($\overrightarrow{C_W}$) and adding it to the correct class HV ($\overrightarrow{C_R}$), as shown in Eq.~\ref{eqn:retrain}.

\begin{equation}
    \begin{aligned}
       \overrightarrow{C_W} = \overrightarrow{C_W} - \overrightarrow{S}, & & \overrightarrow{C_R} = \overrightarrow{C_R} + \overrightarrow{S}
    \end{aligned}
    \label{eqn:retrain}
\end{equation}

\subsection{HDC Inference}
Inference is the process of using unseen data from the inference set to evaluate the trained model's performance. As illustrated in Fig.~\ref{fig:hdc}(c), \model ~calculates the cosine similarity ($\delta$) between the sample HV from the inference set with unknown label (referred to as query HV (qHV, $\overrightarrow{Q_?}$) and each cHV in the AM to obtain the similarity values. The cHV with the most similarity indicates having the most overlap as to the preserved information with the qHV. Therefore, class of the qHV, i.e., the class of the inference sample, is subsequently predicted as $x$ as described in ~\ref{eqn:sim}.

\begin{equation}
    \begin{aligned}
    x = argmax(\delta(\overrightarrow{Q_?}, \mathbb{C}))
    \end{aligned}
    \label{eqn:sim}
\end{equation}

\section{Experimental Results}
\label{sec:results}
\subsection{Experimental Setup}
\label{sec:setup}
\mypara{Datasets} We use 29 binary classification tasks in total from 3 datasets in the popular \textbf{MoleculeNet} benchmark suite for molecule machine learning~\cite{wu2018moleculenet}. For each dataset, we perform 0.8/0.2 random, stratified and scaffold split to build our training and inference set and repeat 5 experiments to get average performance with error bars. Details of the datasets are as follows:
\begin{itemize}
    \item \textbf{BBBP}~\cite{martins2012bayesian} contains 2052 drug compounds and their binary label (positive or negative) of permeability to the blood-brain barrier.
    \item \textbf{Clintox}~\cite{gayvert2016data} contains 1491 drug compounds and their binary label (positive or negative) of 1) clinical trial toxicity and 2) FDA approval status.
    \item \textbf{SIDER}~\cite{kuhn2016sider} contains 1428 marketed drugs and their adverse drug reactions (ADR) in 27 individual tasks per \textbf{MedDRA} classifications~\cite{brown1999medical}. Each task aims to classifying the positive (active) or negative (inactive) relationship between the drug compound and the ADR disorders of system organs. 
\end{itemize}

\mypara{Baseline Models} We compare \model ~with various baseline methods which are roughly in three categories: traditional learning models, GNNs and RNNs.
\begin{itemize}
    \item \textbf{Traditional learning Models} including logistic regression \textbf{(LR)}, random forest \textbf{(RF)}, and support vector machine \textbf{(SVM)} implemented and reported in the \textit{MoleculeNet} benchmark~\cite{wu2018moleculenet} and \textit{DeepChem}~\cite{ramsundar2019deep} framework.
    \item \textbf{Weave}~\cite{kearnes2016molecular}, which is a graph convolution method that takes both local chemical environment and atom connectivity in featurization.
    \item \textbf{MolCLR}~\cite{wang2021molclr}, which is a GNN with contrastive learning of representations with augmentations of atom masking, bond deletion, and subgraph removal.
    \item \textbf{D-MPNN}~\cite{yang2019learned, swanson2019message}, which is the directed message passing neural network that operates on molecular graphs. 
    \item \textbf{LSTM}~\cite{quan2018system}, which applies a modified version of the Smi2Vec tool to convert SMILE strings into atom vectors and then apply long short term memory (LSTM) RNN for classification.
    \item \textbf{BiGRU}~\cite{lin2020novel}, which also uses Smi2Vec. It leverages the bidirectional gated recurrent unit (BiGRU) RNN to train sample vectors embedded in the atomic matrix.
    \end{itemize}

\subsection{Metrics}
Drug discovery datasets are mostly significantly imbalanced, thus accuracy is generally not considered as a valid metric to reflect performance of a model. Receiver operating characteristics (ROC) curves and ROC Area-under-curve (AUC) scores are mostly embraced as the metric for model prediction performance, as suggested by benchmark datasets along with majority of literature~\cite{wu2018moleculenet, ramsundar2019deep, mayr2018large}. Since HDC models are predicting using similarities, the ``probability'' used in calculating the ROC-AUC score requires specific definition. We propose to use ``confidence level'' (for being positive) $\eta$ in Eq.~\ref{eqn:score}s. Confidence level is derived from similarities between query HV and the class HVs. The larger the difference, the higher the confidence of the HDC model prediction. Because the range of similarity difference is [-2, 2], to perform linear transformation to map the range of confidence level to [0, 1], we accordingly set 1/2 as the average value and 1/4 for coefficient of similarity difference, conforming with the probabilities.

\begin{equation}
    \eta = \frac{1}{2} + \frac{1}{4}(\delta(\overrightarrow{Q_?}, \overrightarrow{C_P}) - \delta(\overrightarrow{Q_?}, \overrightarrow{C_N}))
    \label{eqn:score}
\end{equation}

\begin{table*}[h!]
  \centering
    \begin{tabular}{c|ccc|ccc|ccc}
    \toprule
    split & \multicolumn{3}{c|}{random} & \multicolumn{3}{c|}{stratified} & \multicolumn{3}{c}{scaffold} \\
    dataset & Clintox & BBBP  & SIDER & Clintox & BBBP  & SIDER & Clintox & BBBP  & SIDER \\
    \midrule
    \model & $\mathbf{0.976}^{(1)}$ & $0.879^{(3)}$ & $0.599^{(4)}$ & $0.973^{(2)}$ & $0.916^{(3)}$ & $0.61^{(2)}$ & $\mathbf{0.987}^{(1)}$ & $0.844^{(2)}$ & $0.566^{(4)}$ \\
    tokenization & char  & char  & PE    & char  & PE    & PE    & char  & char  & PE \\
    gram size & trigram & trigram & unigram & trigram & unigram & trigram & bigram & bigram & bigram \\
    \midrule
    LR    & 0.733 & 0.737 & 0.643 & -     & 0.728 & -     & -     & 0.699 & - \\
    RF    & 0.551 & 0.811 & 0.567 & -     & 0.736 & -     & 0.712 & 0.770 & 0.549 \\
    SVM   & 0.669 & 0.67  & \textbf{0.656} & -     & 0.587 & -     & 0.669 & 0.729 & \textbf{0.682} \\
    Weave & 0.948 & 0.832 & 0.581 & -     & -     & -     & 0.823 & 0.837 & 0.543 \\
    MolCLR & -     & -     & -     & -     & -     & -     & 0.932 & 0.736 & 0.68 \\
    D-MPNN & 0.892 & \textbf{0.92}  & 0.639 & 0.898 & 0.932 & \textbf{0.655} & 0.874 & \textbf{0.915} & 0.606 \\
    LSTM  & -     & 0.832 & -     & -     & 0.876 & 0.530 & -     & -     & - \\
    BiGRU & -     & 0.889 & -     & \textbf{0.978} & \textbf{0.946} & 0.607 & -     & -     & - \\
    \bottomrule
    \end{tabular}%
      \caption{\model ~vs. Baselines on 3 datasets by average ROC-AUC score. Bold: the highest score. ``-'': data unavailable. Superscript ``(k)'': \model ~ranks $k$-th place amongst all the available models under current dataset and split method.}
  \label{tab:beyond}%
\end{table*}%

The experimental results are presented in two parts: the comparison between \model ~and other baseline models as well as the comparison within \model ~configurations.

\subsection{\model ~vs. Baselines}

Most of the baseline models report results not as exhaustive as \model ~in terms of split strategy. Therefore, we are performing comparison by ``best effort'', i.e., we compare best performing \model ~with the baseline with data available under each split method of all the tasks. We decide not to report error bars or variations for this comparison because 1). baseline models may also use different numbers of runs for average and/or cross-validations, and 2). some of the baselines just simply did not report the error bars or variation. However, for all the \model ~versions we implemented, we report all the error bars in Table~\ref{tab:pe} and Table~\ref{tab:char}.

We can observe from the results at Table~\ref{tab:beyond} that, in general, \model ~is achieving high ROC-AUC scores across datasets. For each split, \model ~achieves a dataset-average ROC-AUC scores of \textbf{0.818}, \textbf{0.833} and \textbf{0.799} respectively, \textbf{ranking first on random and scaffold split and second on stratified split}, amongst the models with data available. Particularly, \model ~performs greatly on the Clintox dataset particularly with scaffold split which are often regarded more challenging than the other splits where most of other baseline models are suffering from degradation, the score of \model ~even increases instead.

We have an interesting observation that traditional models such as LR, RF and SVM exhibit poor performance over Clintox and BBBP datasets with significantly low score, however, they show competitive score for the SIDER dataset. On the contrary, while NNs usually performs greatly on Clintox and BBBP datasets, they only show sub-par score even lower than some traditional models, e.g., SVM achieves highest accuracy on SIDER dataset with random and scaffold split.

Robustness-wise, \model ~also outperforms other models. For example, some GNNs are able to achieve top score on a specific dataset, however, they present much lower score on other datasets. For example, for \textbf{D-MPNN}, although it shows high scores at the BBBP dataset by ranking first at random and scaffold split, its score on Clintox seems mediocre. For Weave, it shows significantly degraded score on Clintox dataset from random split to scaffold split. Such variation on performance would arouse questions on those models' transferability, while for \model, the performance is largely consistent across different datasets and split methods.

\begin{table*}[t]
\vspace{-.5em}
  \centering
    \begin{tabular}{c|ccc|ccc|ccc}
    \toprule
    split & \multicolumn{3}{c|}{random} & \multicolumn{3}{c|}{stratified} & \multicolumn{3}{c}{scaffold}  \\
    gram  & {uni-gram} & {bi-gram} & {tri-gram} & {uni-gram} & {bi-gram} & {tri-gram} & {uni-gram} & {bi-gram} & {tri-gram} \\
    \midrule
    Clintox & {$0.941^{0.010}_{0.015}$} & {$0.897^{0.013}_{0.012}$} & {$0.881^{0.024}_{0.013}$} & {$0.960^{0.014}_{0.024}$} & {$0.970^{0.008}_{0.013}$} & {$0.932^{0.025}_{0.018}$} & {$0.952^{0.009}_{0.014}$} & {$0.966^{0.009}_{0.014}$} & {$0.930^{0.020}_{0.021}$} \\
    BBBP  & {$0.886^{0.014}_{0.024}$} & {$0.875^{0.012}_{0.021}$} & {$0.834^{0.021}_{0.021}$} & {$0.916^{0.014}_{0.014}$} & {$0.908^{0.016}_{0.025}$} & {$0.884^{0.016}_{0.026}$} & {$0.785^{0.024}_{0.023}$} & {$0.802^{0.021}_{0.021}$} & {$0.801^{0.020}_{0.018}$} \\
    SIDER & 0.599 & 0.588 & 0.574 & 0.584 & 0.594 & 0.610 & 0.556 & 0.566 & 0.554 \\
    \bottomrule
    \end{tabular}%
 \caption{\model-PE performance on ROC-AUC score comparison on 3 datasets by average. Superscript and subscript refer to the ceiling and floor of errors. For SIDER dataset, the ROC-AUC score is task-average.}
  \label{tab:pe}%
\end{table*}%

\begin{table*}[h!]
\vspace{-.5em}
  \centering
    \begin{tabular}{c|ccc|ccc|ccc}
    \toprule
    split & \multicolumn{3}{c|}{random} & \multicolumn{3}{c|}{stratified} & \multicolumn{3}{c}{scaffold}  \\
    gram  & {uni-gram} & {bi-gram} & {tri-gram} & {uni-gram} & {bi-gram} & {tri-gram} & {uni-gram} & {bi-gram} & {tri-gram} \\
    \midrule
    Clintox & {$0.955^{0.023}_{0.023}$} & {$0.971^{0.015}_{0.016}$} & {$0.976^{0.016}_{0.025}$} & {$0.956^{0.011}_{0.028}$} & {$0.971^{0.019}_{0.020}$} & {$0.973^{0.010}_{0.014}$} & {$0.966^{0.008}_{0.010}$} & {$0.987^{0.001}_{0.001}$} & {$0.982^{0.002}_{0.002}$} \\
    BBBP  & {$0.850^{0.022}_{0.028}$} & {$0.879^{0.034}_{0.029}$} & {$0.879^{0.026}_{0.020}$} & {$0.860^{0.017}_{0.020}$} & {$0.877^{0.014}_{0.013}$} & {$0.865^{0.012}_{0.015}$} & {$0.805^{0.009}_{0.011}$} & {$0.844^{0.006}_{0.010}$} & {$0.828^{0.004}_{0.002}$} \\
    SIDER & 0.580 & 0.544 & 0.525 & 0.578 & 0.544 & 0.514 & 0.553 & 0.541 & 0.565 \\
    \bottomrule
    \end{tabular}%
      \caption{\model-char performance on ROC-AUC score comparison on 3 datasets by average. Superscript and subscript refer to the ceiling and floor of errors. For SIDER dataset, the ROC-AUC score is task-average.}
  \label{tab:char}%
\end{table*}%

\subsection{Results between different \model ~versions}
In addition to comparing with baseline models, we also evaluate an intensive set of \model ~and dataset configurations, including: two tokenization schemes (\model-PE and \model-char), two gram sizes (uni-gram and bi-gram), and three dataset split methods (random, stratified and scaffold split). 

In general, the performance of \model ~is overall consistent, thus, there is no single configuration that can dominate other configurations for most, if not all, the datasets and split methods. However, we do observe that for the scaffold split, the score variation is generally smaller than that of random and stratified split, as suggested by the error bars.                                                                                                                                                                      
\subsection{Ultra-Low-Cost Computing of \model}
We elaborate the computing cost of \model ~and compare it with SOTA neural networks. 
1) Unlike GNNs, \model ~does not require specific additional effort on pre-training the model. 2) For all the reported datasets together, \model ~is able to achieve the reported accuracy within 10 minutes using CPU only from the commodity desktop (AMD Ryzen\texttrademark ~5 3600 3.6 GHz). Note that this includes both training and inference for each dataset. 3) \model ~also requires less space for model storage as for one binary classification task, model size of \model ~is only around 80kB and during run-time, the memory footprint of \model ~is also generally less than 10MB. For GNNs as a comparison, extensive pre-training can be necessary, e.g., MolCLR requires around 5 days of pre-training using Nvidia\textregistered ~Quadro RTX\texttrademark ~6000 as reported in the corresponding literature~\cite{wang2021molclr}. Neural network models such as GNNs and RNNs are also harder to implement considering the effort of establish multiple layers with considerable amount of nodes with the model size at 100MB level, especially considering the necessity of performing back-propagation during training with millions of parameters in total~\cite{ma2020multi}.

\section{Conclusion}

In this paper, we propose \model, an ultra-low-cost learning model which leverages the novel brain-inspired hyperdimensional computing for molecule property prediction in drug discovery. \model ~projects SMILES strings of drug compound into hypervectors in the hyperdimensional space to extract features. The hypervectors are then used during training, retraining and inference of the HDC model to perform learning tasks. We evaluate \model ~on 29 classification tasks from 3 widely-used benchmark datasets and compare \model ~performance with 8 baseline machine learning models including SOTA GNNs and RNNS. According to experimental results, \model ~is able to achieve highest ROC-AUC score on random and scaffold splits on average across 3 datasets and achieve second-highest on stratified split. Compared with traditional models and NNs, \model ~also requires less training efforts, smaller model size, as well as smaller computation costs. This work marks the potential of using hyperdimensional computing as an alternative to the existing models in the drug discovery domain.
\bibliography{acmart.bib}
\end{document}